\documentclass[conference]{IEEEtran}
\usepackage{cite}
\usepackage{amsmath,amssymb,amsfonts}
\usepackage{algorithmic}
\usepackage{graphicx}
\usepackage{textcomp}
\usepackage[table,xcdraw]{xcolor}
\usepackage{booktabs}
\usepackage{multirow}
\usepackage{hyperref}
\usepackage{svg}
\def\BibTeX{{\rm B\kern-.05em{\sc i\kern-.025em b}\kern-.08em
    T\kern-.1667em\lower.7ex\hbox{E}\kern-.125emX}}

\begin{document}
\title{Fast ML-driven Analog Circuit Layout using Reinforcement Learning and Steiner Trees\\
\thanks{This work has been developed in the project HoLoDEC (project label 16ME0696) which is partly funded within the Research Programme ICT 2020 by the German Federal Ministry of Education and Research (BMBF) and partially supported by the PNRR project iNEST (Interconnected North-Est Innovation Ecosystem) funded by the European Union Next-GenerationEU (Piano Nazionale di Ripresa e Resilienza (PNRR) – Missione 4 Componente 2, Investimento 1.5 – D.D. 1058 23/06/2022, ECS\_00000043).}
}
\IEEEoverridecommandlockouts

% \IEEEpubid{\makebox[\columnwidth]{979-8-3503-3265-0/23/\$31.00~\copyright2024 IEEE \hfill} \hspace{\columnsep}\makebox[\columnwidth]{ }}
\author{
\IEEEauthorblockN{1\textsuperscript{st}
Davide Basso
2\textsuperscript{nd}
Luca Bortolussi
}
\IEEEauthorblockA{\textit{University of Trieste}\\
Trieste, Italy\\
davide.basso@phd.units.it,
lbortolussi@units.it
}
\and
\IEEEauthorblockN{3\textsuperscript{rd}
Mirjana Videnovic-Misic
\IEEEauthorblockA{\textit{Infineon Technologies Austria AT}\\
Villach, Austria\\
mirjana.videnovic-misic@infineon.com}}
\and
\IEEEauthorblockN{4\textsuperscript{th}
Husni Habal
\IEEEauthorblockA{\textit{Infineon Technologies AG}\\
Munich, Germany\\
husni.habal@infineon.com}}
}

% Department of Mathematics, Informatics and Geosciences, University of Trieste
%
\maketitle\IEEEpubidadjcol
%
% make the title area
\hbadness=10000\maketitle\hbadness=1000

\makeatletter
\newcommand{\linebreakand}{%
  \end{@IEEEauthorhalign}
  \hfill\mbox{}\par
  \mbox{}\hfill\begin{@IEEEauthorhalign}
}
\makeatother
\bstctlcite{IEEEexample:BSTcontrol}
\maketitle

\begin{abstract}
This paper presents an artificial intelligence driven methodology to reduce the bottleneck often encountered in the analog ICs layout phase. We frame the floorplanning problem as a Markov Decision Process and leverage reinforcement learning for automatic placement generation under established topological constraints. Consequently, we introduce Steiner tree-based methods for the global routing step and generate guiding paths to be used to connect every circuit block.
Finally, by integrating these solutions into a procedural generation framework, we present a unified pipeline that bridges the divide between circuit design and verification steps. Experimental results demonstrate the efficacy in generating complete layouts, eventually reducing runtimes to $1.5\%$ compared to manual efforts.
\end{abstract}


\section{Introduction}
The layout of analog ICs, traditionally dependent on manual expertise, has been slow to adopt artificial intelligence (AI) advancements that have transformed digital counterpart through Electronic Design Automation (EDA) especially owing to high susceptibility to noise, variations in process, voltage, and temperature. These properties, in fact, translate into several topological requirements that must be met to produce robust layouts, tackling possible parasitics and routability issues.

Analog floorplanning has traditionally required a significant amount of expert knowledge and involved a high degree of repetitive work.
Metaheuristics like simulated annealing (SA), particle swarm optimization (PSO), and genetic algorithms (GA) \cite{MetaheurReview} have been employed to streamline this process, unfortunately lacking of any possibility to reuse past experience information. 
Recently, learning-based techniques as reinforcement learning (RL) have gained traction thanks to their effectiveness on solving combinatorial problems \cite{RL4CO}, to which floorplanning belongs. Very few attempts have been made following this direction, especially in the analog setting \cite{gusmao} and, to our knowledge, the use of RL is still to be explored.
Efforts to automate routing with deep learning (DL) models \cite{DLGlobalRouting} have seen limited real-world application as well, largely due to the scarcity of diverse, public datasets.

We therefore propose two novel approaches for optimal floorplan generation, framing the problem as a Markov Decision Process (MDP) \cite{MDP} and eventually employing RL techniques alone and in cooperation with SA. The floorplan is mathematically described through a topological representation called sequence pair (SP) \cite{SP}. Our models are trained on synthetic floorplans for optimal topological constraint handling, ensuring they generalize well to new, real-world circuits.
Moreover, we introduce an Obstacle Avoiding Rectilinear Steiner Tree (OARSMT) based algorithm, inspired by \cite{OARSMT}, to generate routing guiding paths with optimal nets and metal layer parameters selection. The complete pipeline, integrating both automatic floorplan and routing strategies with a preliminary circuit functional blocks recognition \cite{SRtool}, is illustrated in Figure \ref{fig:flow}. 
\begin{figure}
    \centering
    \includegraphics[width=\columnwidth]{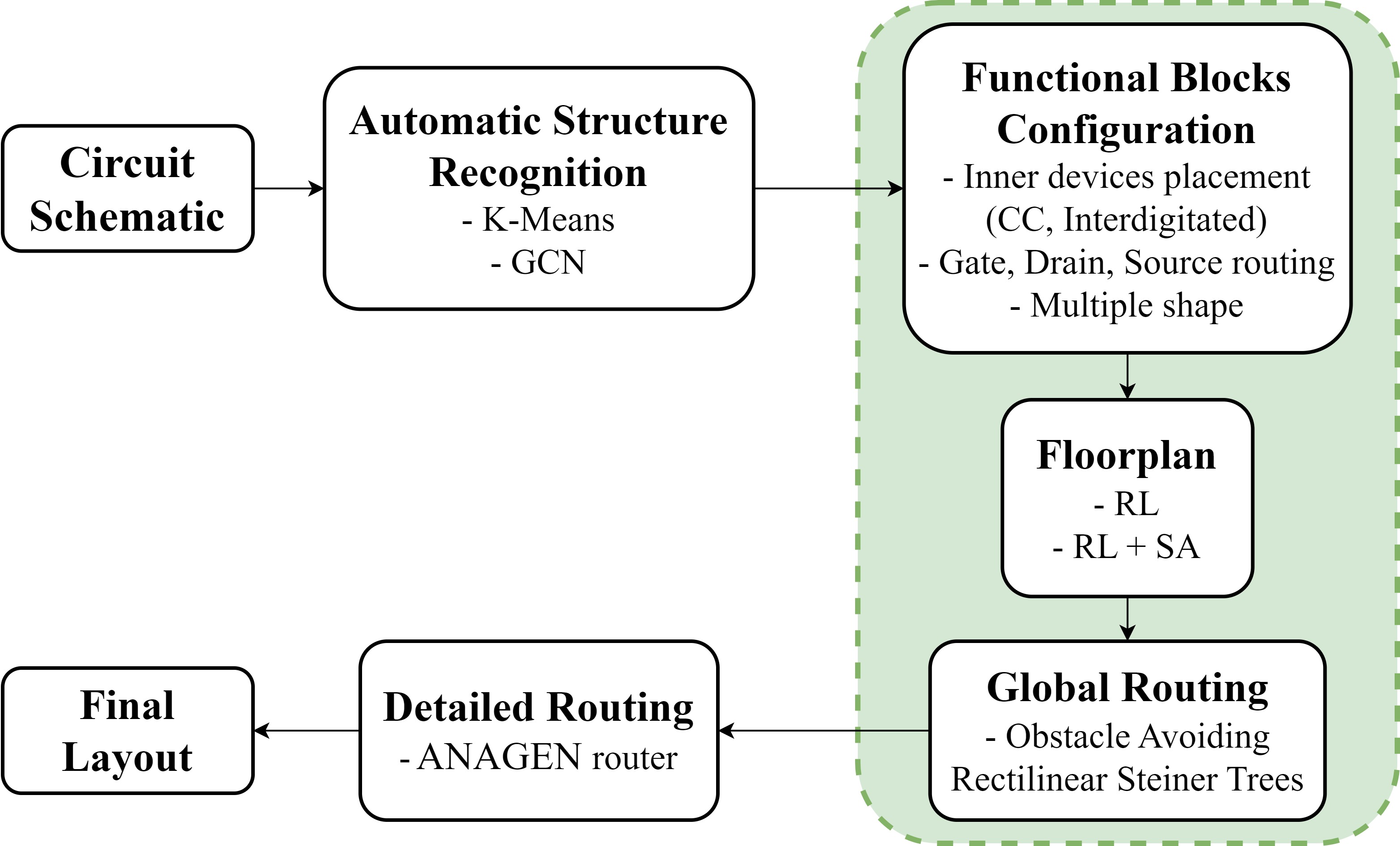}
    \caption{High-level schematic of the AI-powered automated layout pipeline with this paper contributions highlighted in green.}
    \label{fig:flow}
\end{figure}
The key contributions of this work are as follows:
\begin{itemize}
    \item We propose two RL-based automated floorplan generators for analog ICs capable of optimizing circuit area occupation and estimated wirelength. The support of fundamental topological requirements such as device symmetry, alignment and optimal routability is guaranteed along with the possibility to define specific aspect ratios of the final floorplan and a variable shape for each device.
    \item An OARSMT method is devised for global routing, offering immediate guidance for ANAGEN \cite{anagen, anagen2} to complete detailed inter-block connections.
    \item We integrate both techniques into a procedural generator, providing engineers with an automated layout template pipeline. Industrial scenario tests show our method rivals manual layouts in performance, reducing early template generation time from 16 hours to just 57.48 seconds.
\end{itemize}

In the next Section, a brief overview of floorplanning and routing problems is given. Then, in Section \ref{sec:methodology}, the RL-based floorplan generation, OARSMT global routing and their addition to the ANAGEN flow is detailed. Eventually, Section \ref{sec:results} presents results obtained on an Infineon developed OTA circuit and, in Section \ref{sec:conclusions}, conclusions are addressed.

\section{Problem Definition}
\subsection{Floorplanning} 
The goal of floorplanning can be identified as optimizing a predefined cost function encompassing different objectives and constraints. In this study, we focus on pure area minimization of a floorplan $F$ (\ref{area_cost}) or its combination with half-perimeter wirelength (HPWL) within a fixed-outline constraint (\ref{multiple_cost}).
\begin{equation} 
cost = F_{area} \label{area_cost}     
\end{equation} 
\begin{equation} 
cost = \alpha \frac{F_{area}}{\sum_{i=1}^m A_i} + \beta \frac{\text{HPWL}}{\text{HPWL\textsubscript{avg}}}  + (1-\alpha-\beta)(R^* - R)^2 \label{multiple_cost}
\end{equation}
Here, $\alpha$ and $\beta$ are weights in $[0,1]$ balancing area and wirelength terms importance. $A_i$ represents the area of the $i^{th}$ device and $\text{HPWL}_{\text{avg}}$ the average HPWL from the last $100$ simulations, both used for standardization purposes. Finally, $R^*$ and $R$ are respectively the target and current floorplan aspect ratios\looseness=-1. 

\subsection{Routing}
The global routing phase aims to optimize the allocation of on-chip routing resources, commonly discretizing the layout into a grid, to interconnect circuit components given their placement. As per ANAGEN's rules, we treat each circuit block not belonging to the network of interest as an obstacle. Then, a minimal wirelength routing tree connecting all devices, potentially using additional nodes known as Steiner points, can be constructed using rectilinear lines. These paths form the basis for the detailed routing stage, which precisely defines physical interconnections of the final layout.

\section{Automatic Layout Generation}
\label{sec:methodology}
\subsection{Topological Representation of a Circuit Floorplan}
A floorplan can be defined through a topological representation mapping the geometrical relationships of each device. 
In our experiments, a floorplan is represented by a sequence pair, ($\Gamma_1,\Gamma_2$), consisting of two sequences of module identifiers. The relative position of each identifier in both sequences determines the modules' relative spatial arrangement as follows:
\begin{itemize}
    \item Module $i$ is left to module $j$ if $j$ is after $i$ in both $\Gamma_1,\Gamma_2$.
    \item Module $i$ is below to module $j$ if $j$ is before $i$ in  $\Gamma_1$ and after $i$ in $\Gamma_2$.
\end{itemize}
The relations ``right" and ``above" are defined symmetrically, swapping $i$ and $j$.
% The right and above to relationship can be easily retrieved by swapping modules $i$ and $j$ in the above definitions.

\subsection{Simulated Annealing and Reinforcement Learning}
SA is a meta-heuristic that iteratively searches for a function's global optimum from an initial state, in this context a floorplan encoded as a SP, through probabilistic perturbations. Each change is evaluated against a cost function, with the temperature parameter guiding the likelihood of accepting suboptimal moves to escape local minima. During this process, the floorplan is modified by swapping modules, rotating, or reshaping them, while maintaining constraints like symmetry and alignment, as in \cite{symm}. 
On the other hand, RL employs an MDP framework, where an agent discovers optimal solutions by interacting with the environment. Actions $\mathcal{A}$ in RL transition the floorplan between states $\mathcal{S}$. The agent is directed by rewards $\mathcal{R}$, which evaluate the actions' impact, and a discount factor $\gamma$, balancing the importance of immediate versus future rewards. Through this, the agent learns the policy that maximizes the expected sum of rewards.

\subsection{Structures Recognition and Device Shapes Generation}
Given an input analog circuit schematic, we search for an optimal placement of each device minimizing one of the defined cost functions (\ref{area_cost}, \ref{multiple_cost}). 
Utilizing Infineon's structure recognition tool, we apply clustering and graph convolutional networks (GCN) to detect functional blocks within the schematic.
Subsequently, various block shape configurations are generated, ensuring compliance w.r.t. a fixed total device width $W_{\text{tot}} = w_f \times N_f \times M$, respectively being finger width, number of fingers, and device multiplicity, while considering design rule checks and engineering specifications to define permissible parameter ranges. 
Leveraging ANAGEN's capabilities, we tailored intra-block properties to the type of structures identified. For instance, when a differential pair is detected, devices are arranged in an interdigitated or common centroid layout to reduce mismatch, with the latter approach inspired by \cite{Common_centroid}. Moreover, through a reverse engineering process, conventional routing patterns such as metal layer selection, gate, drain and source connections, dummy transistors integration, among other enhancements have been adopted. 

\begin{figure*}[t!]
    \begin{minipage}[t]{0.39\textwidth}
        \centering
        \includegraphics[width=.95\linewidth]{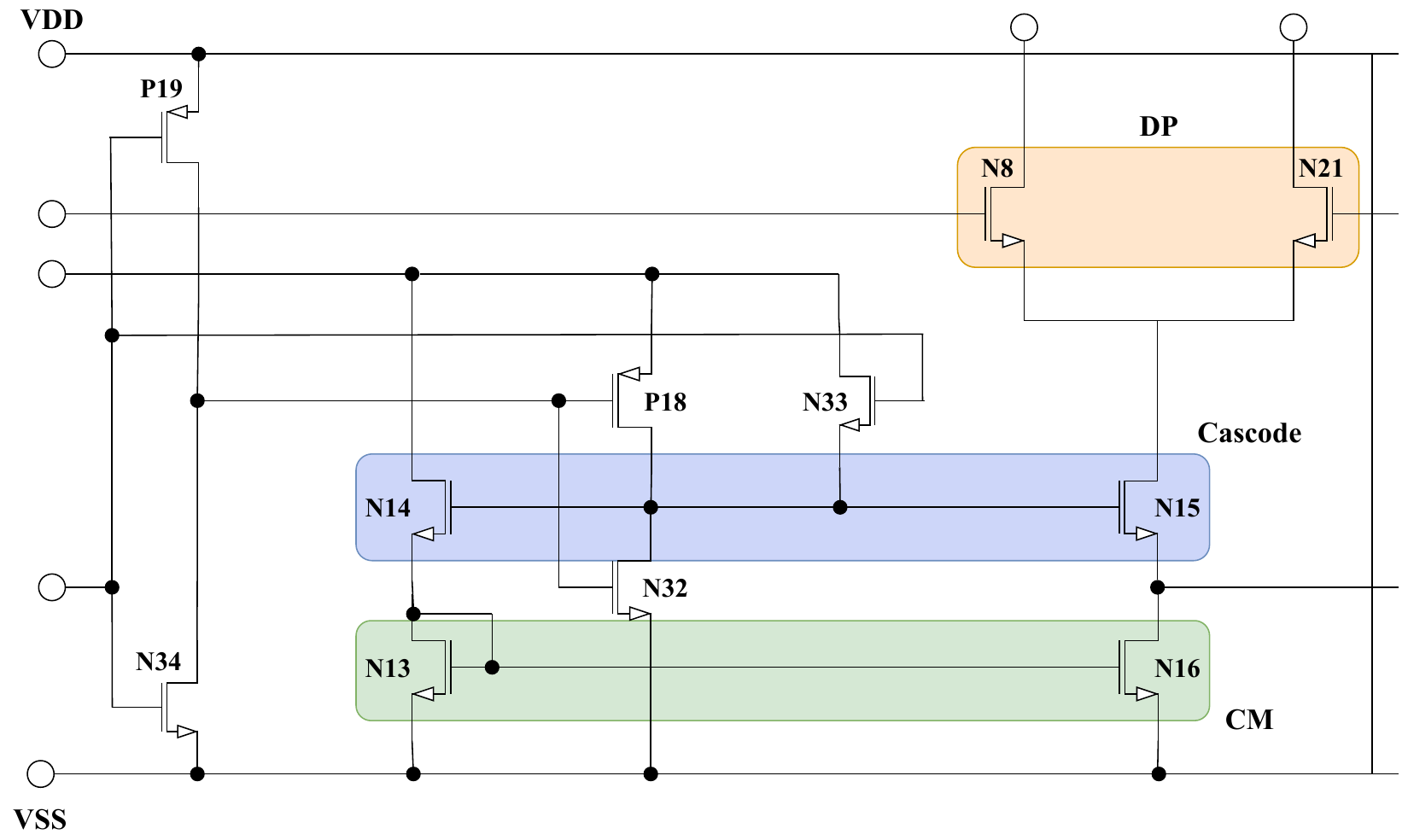} \\ (a)
        \label{fig:ota_schematic}
    \end{minipage}\hfill
    \begin{minipage}[t]{0.28\textwidth}
      \centering
      \includegraphics[width=1.02\linewidth]{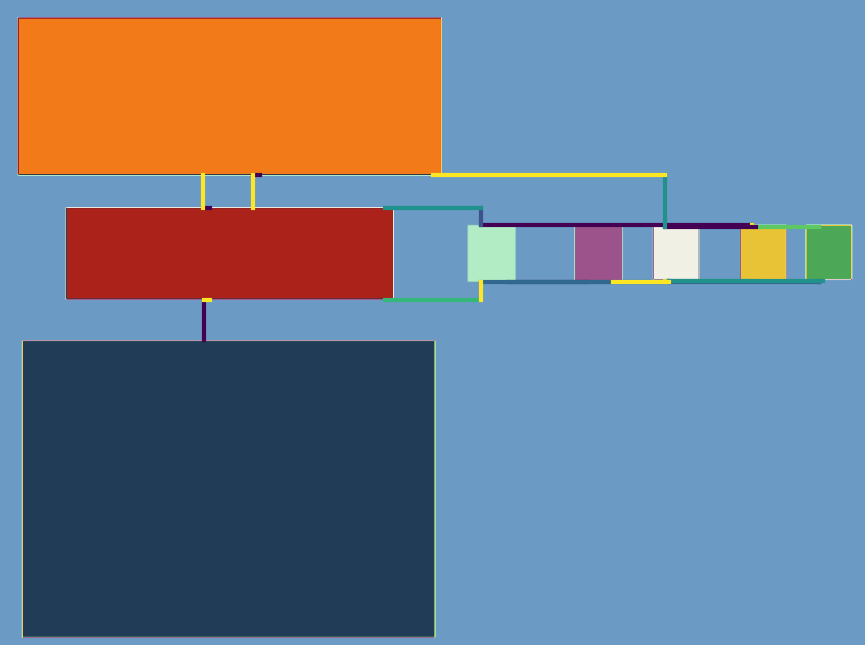}\\ (b)
      \label{fig:fp_routing_RLSA}
     \end{minipage}\hfill
    \begin{minipage}[t]{0.3\textwidth}
      \centering
      \includegraphics[width=.82\linewidth]{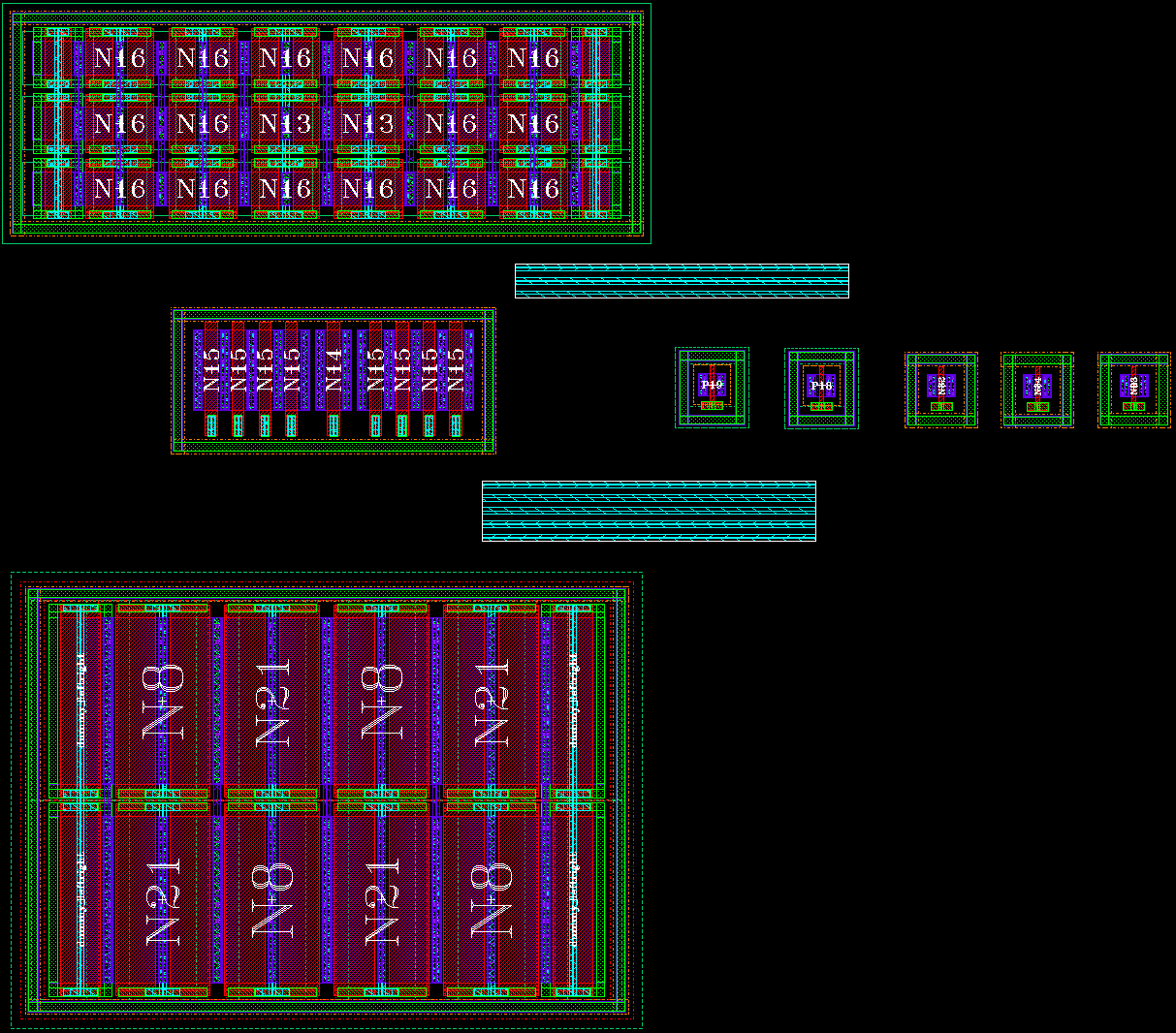}\\ (c)
      \label{fig:virtuoso_original}
   \end{minipage}
    \caption{OTA schematic (a), corresponding floorplan and global routing configuration (b) and its rendering with ANAGEN tracklines (c).}
    \label{fig:schematic_ota_fp_routing}
\end{figure*}

\subsection{RL-SA Floorplanning}
In our first floorplanner setting, an RL agent generates a starting point to be refined by SA. The agent in fact perturbs few times the SP representing the floorplan, as well as doing rotation or altering blocks' shape, thus modifying their parameter setup. During this stage, the agent receives intermediate rewards $cost(s') - cost(s)$, with $s$ and $s'$ denoting the states previous and after agent’s action.
SA then iterates on this early configuration to eventually find an optimal layout. The agent is informed of the goodness of its initialization through a global reward  $cost(s_{m}) - cost(s_{j})$, with $s_{j}$ and $s_{m}$ respectively being post-RL and post-SA state. This strategy leverages RL's capacity for broad search space exploration, compensating for SA's susceptibility to poor initialization that may result in suboptimal layouts.
Moreover, we adopted a cyclical RL-SA collaboration during training, where environment's state is not reset between episodes, i.e. a floorplanning run, allowing the agent to start a new episode upon the SA optimized floorplan state rather than from scratch. This technique, inspired by Mirhoseini et al. \cite{cylcicRLSA}, proved to enhance agent's exploration efficiency, leading to optimized solutions with smaller costs in shorter runtimes\looseness=-1.

\subsection{Pure RL Floorplanning}
An RL agent is now in charge of finding optimal searching techniques to generate a floorplan entirely on its own. The search phase entails sampling a neighboring state using the same set of actions described for SA. The agent has then to decide whether to accept or not the new supplied sample. 
In order to better direct the agent through its exploration and somehow mimic the SA search behaviour, the environment state is augmented with the following additional metrics:
\begin{itemize}
    \item Cost related: The agent keeps track of the current, minimum, average, and neighboring state costs to make better informed decisions; for example, it will likely reject states if their cost substantially exceeds the observed minimum\looseness=-1.
    \item Search phase related: A scalar, analogous to the temperature parameter in SA, represents the current optimization step, informing at state-level the RL agent on its phase in the optimization process.
\end{itemize}
Being solely dependent on the number of circuit devices, this design enhances model capability to learn from disparate floorplan configuration and ensures its reusability across layouts with diverse user-defined constraints during inference.

\subsection{RL Training Setup}
In our study, Proximal Policy Optimization (PPO) \cite{PPO} state-of-the-art actor-critic method is employed for agent training. 
To promote generalization, robustness, and following empirical trials, each model is trained on a diverse set of synthetic floorplans featuring varying topological constraints and aspect ratios. We train one model for each circuit category, differentiated by the count of devices, spanning from 5 to 20.
The RL-SA framework is trained using $128$ RL steps and $10$ epochs, with SA performing $2000$ steps starting at a temperature of $15$. The standalone RL setup undertakes a longer training with $5000$ steps and $50$ epochs for thorough learning. Both frameworks' neural networks use $3$ and $2$ hidden layers with $128$ neurons for the actor and critic, respectively. Average training times for RL-SA is 5 minutes, while for pure RL it is 10 hours\looseness=-1.

\subsection{Post-processing \& Routing}
The floorplan generated by previous algorithms is refined using a congestion estimation technique \cite{congestion_estimation}. This step involves grid-based analysis of wire capacities to produce a congestion map, which guides the redistribution of circuit blocks for a practical and routable layout.
At this point, we build the OARSMT for each net specified in the netlist while minimizing wirelength and avoiding blockages present in the floorplan. The resulting routing tree is decomposed into horizontal and vertical segments and bundled into conduits where feasible. 
These conduits, containing details on the connected devices, associated net, chosen metal layer, are processed by ANAGEN detailed router to finally connect all circuit's devices\looseness=-1.

\section{Experimental Results and Discussion}
\label{sec:results}

Our automated layout frameworks are developed using Python 3.9 and use the stable-baselines3 library \cite{stable-baselines3} to define the RL environment and PPO model. 
To validate the RL-based floorplanning algorithms, performance metrics on $3$ different circuits are benchmarked against established metaheuristics, including SA, PSO and GA, whose hyperparameters can be seen in Table \ref{tab:meta_hyperparams}. Table \ref{tab:RLVSmeta} data clearly demonstrate the RL methods’ superiority in producing more compacted and wirelength-optimized floorplans. While the full RL approach does incur slightly longer runtimes anyway achieving best packing results, its hybridization with SA strikes a balance between optimal metrics and metaheuristic comparable speed.

\begin{table}[]
\caption{Metaheuristics Hyperparameters}
\label{tab:meta_hyperparams}
\resizebox{\columnwidth}{!}{%
\begin{tabular}{@{}c|cc|ccc|cccc@{}}
\toprule
\textbf{Method} & \multicolumn{2}{c|}{\textbf{SA}} & \multicolumn{3}{c|}{\textbf{GA}} & \multicolumn{4}{c}{\textbf{PSO}} \\ \midrule
\textbf{Hyperparam.} & Temp. & Steps & Mut. & Cross. & Size & Inertia & Cogn. & Soc. & Size \\
\textbf{Value} & 15 & 5000 & 0.1 & 0.9 & 200 & 0.8 & 1.49 & 1.49 & 200 \\ \bottomrule
\end{tabular}%
}
\end{table}

\begin{table}[]
\centering
\caption{Runtime, Floorplan Empty Space, and HPWL comparison between RL-Based Algorithms and Traditional Metaheuristics.}
\label{tab:RLVSmeta}
\resizebox{\columnwidth}{!}{%
\begin{tabular}{ccccccccc}
\hline
\textbf{Circuit}        & \textbf{\# Devices} & \textbf{Algorithm}          & \multicolumn{2}{c}{\textbf{Runtime (s)}}                                      & \multicolumn{2}{c}{\textbf{Empty space (\%)}}                         & \multicolumn{2}{c}{\textbf{HPWL (\textmu m)}}                                               \\ \cline{4-9} 
\textbf{}               & \textbf{}             & \textbf{}                   & \textbf{mean}                         & \textbf{std}                          & \textbf{mean}                 & \textbf{std}                          & \textbf{mean}                          & \textbf{std}                           \\ \hline
                        &                       & RL                          & 25.70                                 & 0.31                                  & \textbf{12.14}                & 3.85                                  & 73.57                                  & 7.31                                   \\
                        &                       & RL-SA                       & 3.60                                  & 2.54                                  & 14.02                         & 5.24                                  & 72.44                                  & 10.16                                  \\
                        &                       & \cellcolor[HTML]{C0C0C0}SA  & \cellcolor[HTML]{C0C0C0}\textbf{2.69} & \cellcolor[HTML]{C0C0C0}3.28          & \cellcolor[HTML]{C0C0C0}16.58 & \cellcolor[HTML]{C0C0C0}5.08          & \cellcolor[HTML]{C0C0C0}75.60          & \cellcolor[HTML]{C0C0C0}8.93           \\
                        &                       & \cellcolor[HTML]{C0C0C0}GA  & \cellcolor[HTML]{C0C0C0}4.54          & \cellcolor[HTML]{C0C0C0}0.20          & \cellcolor[HTML]{C0C0C0}26.94 & \cellcolor[HTML]{C0C0C0}9.81          & \cellcolor[HTML]{C0C0C0}87.60          & \cellcolor[HTML]{C0C0C0}16.76          \\
\multirow{-5}{*}{OTA-1} & \multirow{-5}{*}{5}   & \cellcolor[HTML]{C0C0C0}PSO & \cellcolor[HTML]{C0C0C0}4.06          & \cellcolor[HTML]{C0C0C0}\textbf{0.03} & \cellcolor[HTML]{C0C0C0}12.85 & \cellcolor[HTML]{C0C0C0}\textbf{3.28} & \cellcolor[HTML]{C0C0C0}\textbf{69.72} & \cellcolor[HTML]{C0C0C0}\textbf{5.78}  \\ \hline
                        &                       & RL                          & 35.86                                 & 0.38                                  & \textbf{10.19}                & 4.18                                  & 135.06                                 & 24.71                                  \\
                        &                       & RL-SA                       & 3.42                                  & 1.12                                  & 13.61                         & 5.53                                  & \textbf{127.16}                        & 17.32                                  \\
                        &                       & \cellcolor[HTML]{C0C0C0}SA  & \cellcolor[HTML]{C0C0C0}\textbf{1.76} & \cellcolor[HTML]{C0C0C0}1.63          & \cellcolor[HTML]{C0C0C0}14.38 & \cellcolor[HTML]{C0C0C0}7.14          & \cellcolor[HTML]{C0C0C0}136.09         & \cellcolor[HTML]{C0C0C0}15.05          \\
                        &                       & \cellcolor[HTML]{C0C0C0}GA  & \cellcolor[HTML]{C0C0C0}5.14          & \cellcolor[HTML]{C0C0C0}0.05          & \cellcolor[HTML]{C0C0C0}22.43 & \cellcolor[HTML]{C0C0C0}8.42          & \cellcolor[HTML]{C0C0C0}164.80         & \cellcolor[HTML]{C0C0C0}45.51          \\
\multirow{-5}{*}{OTA-2} & \multirow{-5}{*}{8}   & \cellcolor[HTML]{C0C0C0}PSO & \cellcolor[HTML]{C0C0C0}4.82          & \cellcolor[HTML]{C0C0C0}\textbf{0.04} & \cellcolor[HTML]{C0C0C0}11.75 & \cellcolor[HTML]{C0C0C0}\textbf{3.54} & \cellcolor[HTML]{C0C0C0}137.74         & \cellcolor[HTML]{C0C0C0}\textbf{13.37} \\ \hline
                        &                       & RL                          & 28.74                                 & 0.35                                  & \textbf{14.30}                & 5.19                                  & \textbf{220.50}                        & \textbf{31.84}                         \\
                        &                       & RL-SA                       & \textbf{2.77}                         & 2.35                                  & 14.90                         & 5.34                                  & 249.24                                 & 29.43                                  \\
                        &                       & \cellcolor[HTML]{C0C0C0}SA  & \cellcolor[HTML]{C0C0C0}5.86          & \cellcolor[HTML]{C0C0C0}12.62         & \cellcolor[HTML]{C0C0C0}14.97 & \cellcolor[HTML]{C0C0C0}4.87          & \cellcolor[HTML]{C0C0C0}236.44         & \cellcolor[HTML]{C0C0C0}33.85          \\
                        &                       & \cellcolor[HTML]{C0C0C0}GA  & \cellcolor[HTML]{C0C0C0}5.48          & \cellcolor[HTML]{C0C0C0}0.26          & \cellcolor[HTML]{C0C0C0}28.12 & \cellcolor[HTML]{C0C0C0}11.30         & \cellcolor[HTML]{C0C0C0}320.26         & \cellcolor[HTML]{C0C0C0}69.27          \\
\multirow{-5}{*}{Bias}  & \multirow{-5}{*}{11}  & \cellcolor[HTML]{C0C0C0}PSO & \cellcolor[HTML]{C0C0C0}5.83          & \cellcolor[HTML]{C0C0C0}\textbf{0.11} & \cellcolor[HTML]{C0C0C0}18.51 & \cellcolor[HTML]{C0C0C0}\textbf{4.72} & \cellcolor[HTML]{C0C0C0}311.33         & \cellcolor[HTML]{C0C0C0}35.40          \\ \hline
\end{tabular}%
}
\end{table}

Our methodology is applied to an $11$-device Operational Transconductance Amplifier (OTA), featuring a differential pair, current mirror, and cascode structure, showcased in Fig. \ref{fig:schematic_ota_fp_routing}a for practical evaluation.
Considering the identified functional blocks, vertical and horizontal alignment constraints are imposed to promote layout regularity.
The proposed initial template, shown in Figs. \ref{fig:schematic_ota_fp_routing}b and \ref{fig:schematic_ota_fp_routing}c, is further refined, specifically by shifting certain devices to the left, as depicted in Fig. \ref{fig:layout_comparison}. Quantitative comparisons regarding layout generation times, spatial efficiency and routing wirelength are detailed in Table \ref{tab:autoVSengineer}.
The data underscore the efficacy of the automated process, significantly reducing the time needed to produce fully optimized layout w.r.t. manual methods from $24$ hours to $\backsim 21$ minutes. Moreover, the produced layout is $13.8\%$ more compact, without compromising routing quality.
Manual refinements to the automated layout were needed, largely due to the ongoing development of ANAGEN's router, which is yet to reach its full potential for seamless integration. Nonetheless, the automatic floorplans generation, coupled with strategic routing guidance for placing ANAGEN's tracklines, undeniably alleviates much of the engineers' workload.

\begin{table}[]
\small
\centering
\caption{Comparative Analysis of Layout Generation Between Our Automated Method and Expert.}
\label{tab:autoVSengineer}
\begin{tabular}{@{}lccc@{}}
\toprule
\textbf{Metric} & \textbf{Manual} & \textbf{Automated} & \textbf{\textbf{Reduction}} \\ \midrule
Template Generation (s)         & 57600  & 57.48   & \textbf{99.9\%} \\
Refinement Effort (s)           & 28800  & 1200   & \textbf{95.8\%} \\
Layout Area ($\text{\textmu m}^2$)          & 884.43 & 762.29 & \textbf{13.8\%} \\
Wirelength  (\textmu m)            & 533.47 & 453.52 & \textbf{14.9\%} \\ \bottomrule
\end{tabular}
\end{table}

\begin{figure}[]
    \centering
      \includegraphics[width=.45\columnwidth]{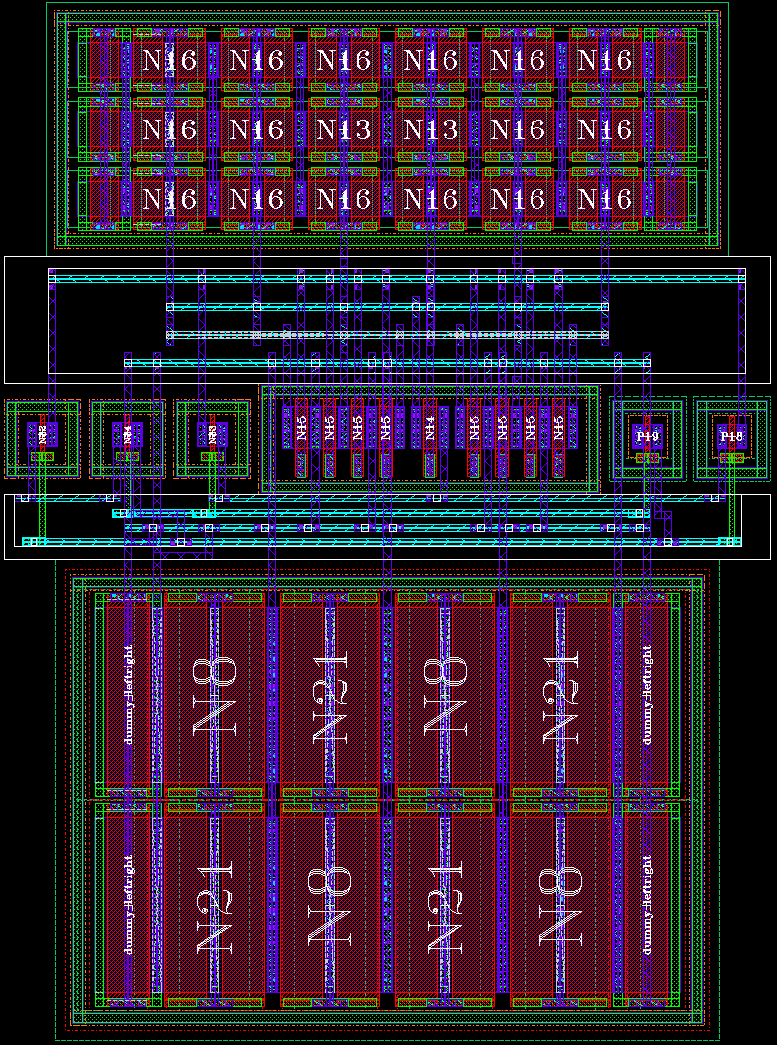}\\ (a)
      \label{fig:ours_ota}

      \vspace{0.15cm}
      
      \includegraphics[trim={0 3.7cm 0 4cm},clip, width=.705\columnwidth]{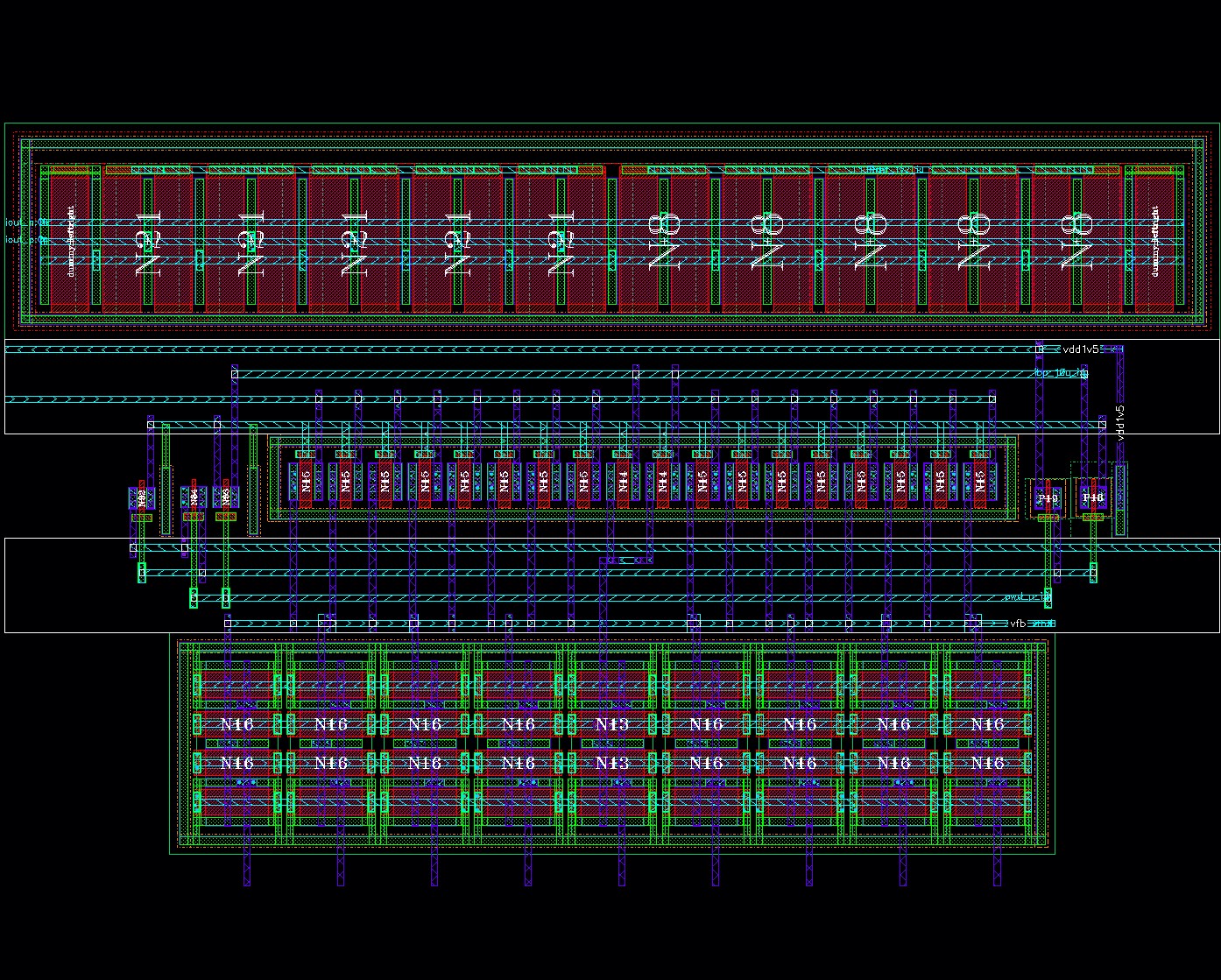}\\ (b)
      \label{fig:gilbert_ota}
    \caption{OTA layout generated from our automatic pipeline (a) and manually from a physical design engineer (b) using ANAGEN layout generator.}
    \label{fig:layout_comparison}
\end{figure}

\section{Conclusions}
\label{sec:conclusions}
In this work, we addressed the complex challenge of analog ICs layout with an AI-driven approach.
First, we detailed two RL-based floorplanning algorithms capable to deal with topological constraints, accommodating variable internal device and functional block configurations while minimizing area and HPWL objectives. 
Additionally, we developed an OARSMT global router, offering valuable support to physical designers during this phase.
Integrated into the ANAGEN framework, our methods have yielded complete layouts in a consistently reduced timeframe, from several hours to just few minutes, shortening time-to-market for analog ICs while adhering to stringent industry quality standards. 

\bibliographystyle{IEEEtran} 
\bibliography{IEEEabrv,refs}

\end{document}